\documentclass[conference]{IEEEtran}    
\usepackage{amsmath,amssymb,epsfig,subcaption,url, nopageno}

\usepackage{graphicx}
\usepackage{amsmath}
\usepackage{amssymb}
\usepackage{booktabs}

\usepackage{times}
\usepackage{epsfig}
\usepackage{graphicx}
\usepackage{xcolor}
\usepackage{caption}
\usepackage{subcaption}
\usepackage{array}

\makeatletter
\newcommand\extralabel[2]{{\edef\@currentlabel{\@currentlabel#2}\label{#1}}}
\makeatother

\usepackage[breaklinks,colorlinks]{hyperref}

\usepackage[capitalize]{cleveref}

\title{Cesium Tiles for High-realism Simulation and Comparing SLAM Results in Corresponding Virtual and Real-world Environments}

\author{
\IEEEauthorblockN{
Chris Beam\IEEEauthorrefmark{1}, 
Jincheng Zhang\IEEEauthorrefmark{1}, 
Nicholas Kakavitsas\IEEEauthorrefmark{2}, 
Collin Hague\IEEEauthorrefmark{2}, 
Artur Wolek\IEEEauthorrefmark{2},
and Andrew Willis\IEEEauthorrefmark{1}}
\IEEEauthorblockA{
\IEEEauthorrefmark{1}
Department of Electrical and Computer Engineering\\
Email: \{cbeam18, jzhang72, awillis\} @charlotte.edu \\
\IEEEauthorblockA{
\IEEEauthorrefmark{2}
Department of Mechanical Engineering and Engineering Science\\
Email: \{nkakavit, chague, awolek\} @charlotte.edu}
University of North Carolina at Charlotte, \\
Charlotte, NC 28223 USA}
}

\date{} 

\begin{document}
\maketitle

\begin{abstract}
  This article discusses the use of a simulated environment to predict algorithm results in the real world. Simulators are crucial in allowing researchers to test algorithms, sensor integration, and navigation systems without deploying expensive hardware. This article examines how the AirSim simulator, Unreal Engine, and Cesium plugin can be used to generate simulated digital twin models of real-world locations. Several technical challenges in completing the analysis are discussed and the technical solutions are detailed in this article. Work investigates how to assess mapping results for a real-life experiment using Cesium Tiles provided by digital twins of the experimental location. This is accompanied by a description of a process for duplicating real-world flights in simulation. The performance of these methods is evaluated by analyzing real-life and experimental image telemetry with the Direct Sparse Odometry (DSO) mapping algorithm. Results indicate that Cesium Tiles environments can provide highly accurate models of ground truth geometry after careful alignment. Further, results from real-life and simulated telemetry analysis indicate that the virtual simulation results accurately predict real-life results. Findings indicate that the algorithm results in real life and in the simulated duplicate exhibited a high degree of similarity. This indicates that the use of Cesium Tiles environments as a virtual digital twin for real-life experiments will provide representative results for such algorithms. The impact of this can be significant, potentially allowing expansive virtual testing of robotic systems at specific deployment locations to develop solutions that are tailored to the environment and potentially outperforming solutions meant to work in completely generic environments.
\end{abstract}


\section{Introduction}

This article discusses how to leverage newly available 3D geospatial technologies to reproduce real-world flight experiments with Unmanned Aerial Systems (UAS)  in simulated environments. Work seeks to compare algorithm results from simulation flights with real-world flights at the same location. Highly realistic recreations of real-world environments significantly benefit UAS development by providing a sophisticated and controlled testing ground for UAS technologies. These environments, often generated through advanced simulation platforms, enable researchers and engineers to replicate complex real-world scenarios with meticulous detail, including urban landscapes, terrain variations, vegetation, and diverse weather conditions. This level of realism facilitates thorough testing and validation of UAS capabilities, such as navigation algorithms, obstacle detection, and collision avoidance systems, in a risk-free virtual environment. Realistic simulations enhance the training of autonomous flight systems by exposing them to a broad spectrum of challenging situations, ultimately accelerating the development, refinement, and validation of UAS technologies before deployment in the actual, and sometimes hazardous, operational environments.


\begin{figure}
    \begin{subfigure}{0.45\linewidth}
        \centering
        \includegraphics[height=2.48cm]{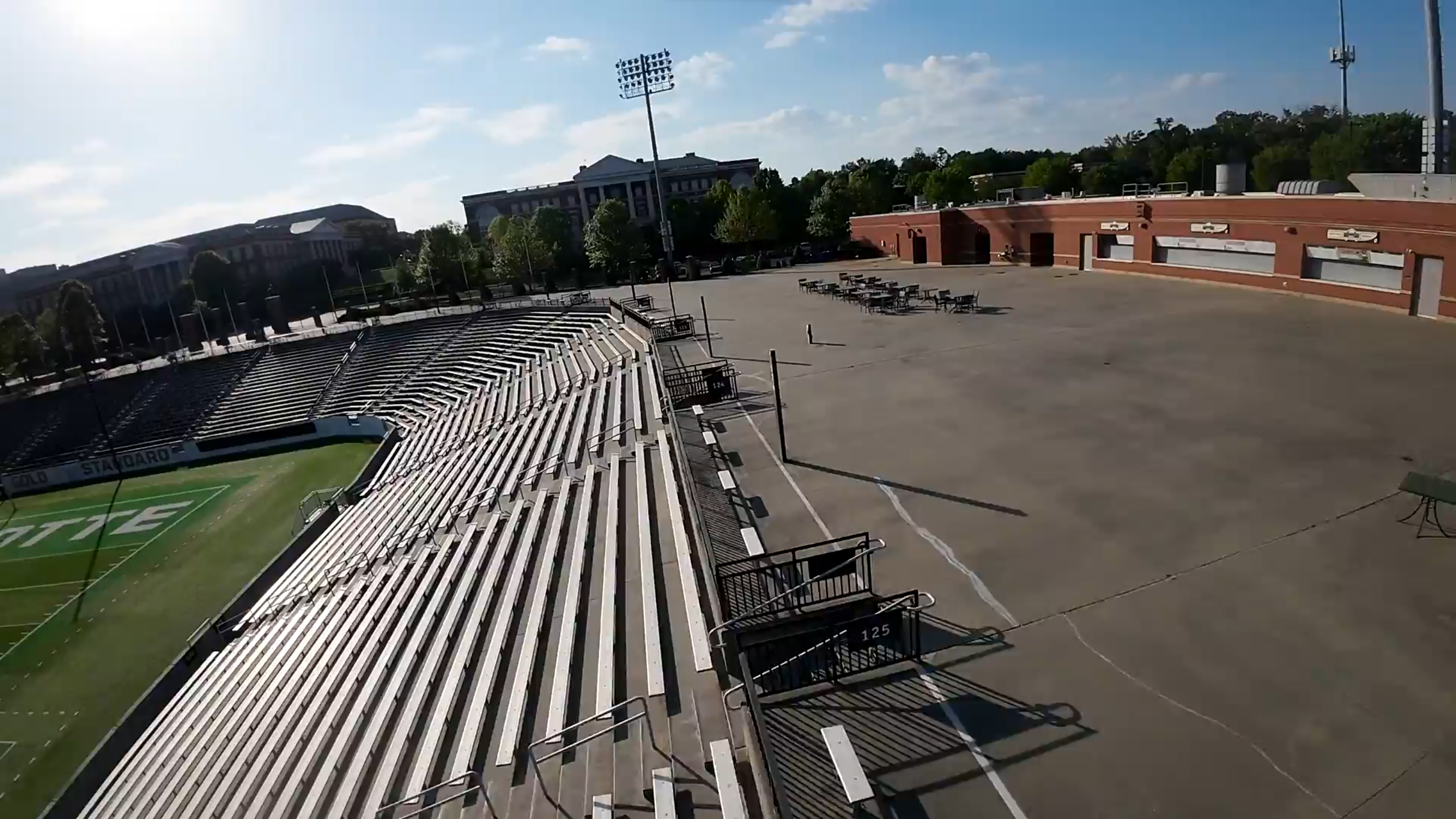}
        \caption{}
        \label{fig:real_world_flight_segment}
    \end{subfigure}
    \begin{subfigure}{0.45\linewidth}
        \centering
        \includegraphics[height=2.48cm]{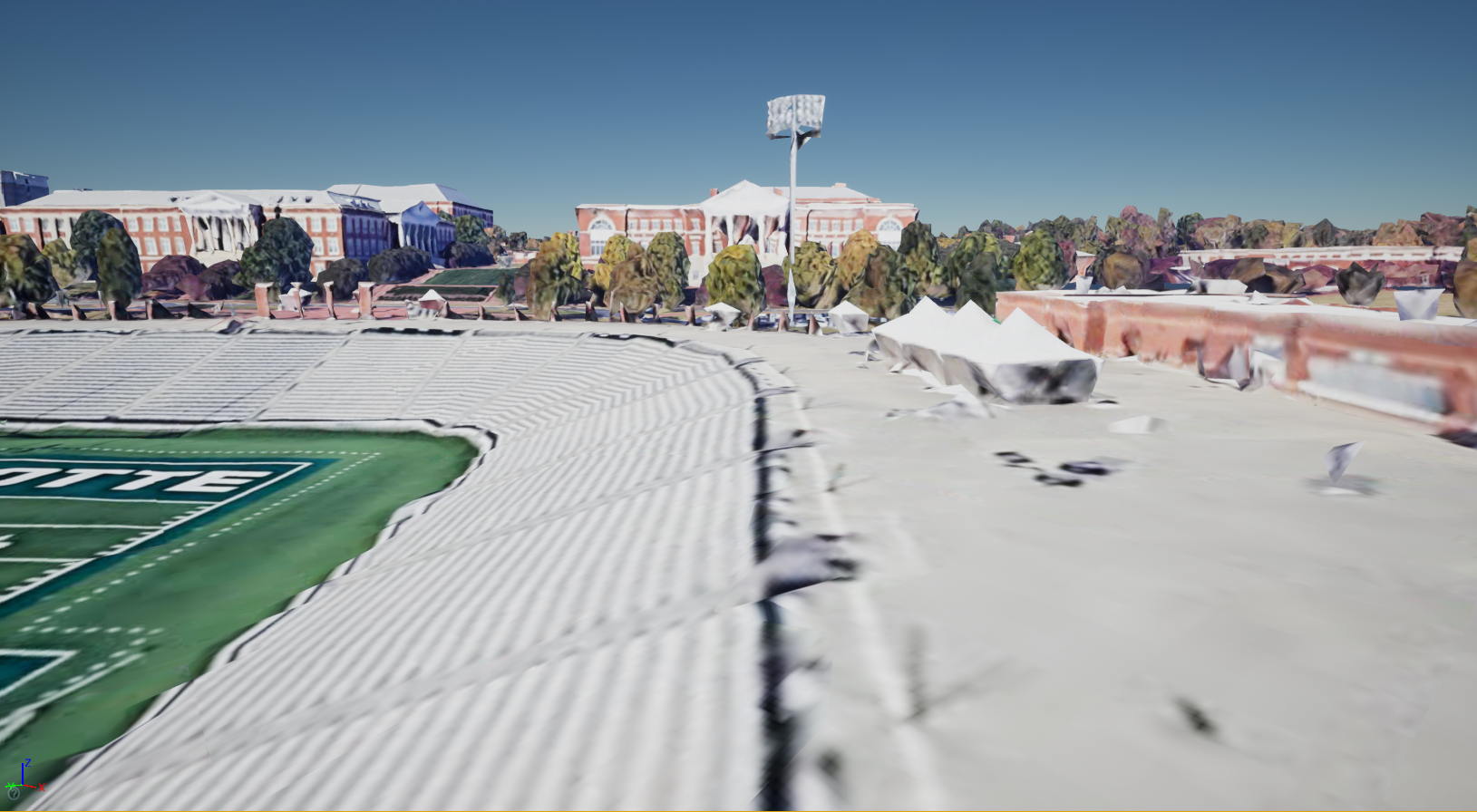}
        \caption{}
        \label{fig:sim_world_flight_segment}
    \end{subfigure}
    \caption{(a,b) show images collected in real-life and simulated aerial experiments over the UNC Charlotte football stadium. This article proposes techniques that use Google's realistic 3D models in UAS 3D mapping and SLAM research.}
\end{figure}
 


Utilizing platforms like Google Maps for simulated environments enhances the precision and richness of UAS testing and mapping endeavors. Google Maps, renowned for its extensive real-world geographic data, serves as a foundational tool to replicate and simulate diverse landscapes within UAS development. By harnessing the intricacies captured in Google Maps—ranging from urban cityscapes to remote terrains—developers can construct highly detailed virtual environments. These environments, driven by Google Maps' comprehensive data, offer an unparalleled level of realism, enabling UAS to navigate and interact within simulated settings mirroring real-world scenarios. Leveraging Google Maps' vast repository, developers can extrapolate detailed geographic information, enriching the maps used as priors for UAS deployment. This integration not only fosters precise testing but also facilitates the creation of robust map priors that closely align with the complexities of actual operational environments, enhancing the efficiency and adaptability of UAS upon real-world deployment.

The Cesium Tiles framework \cite{cesiumtiles} provides access to the highly accurate 3D geometric models developed by Google which previously were only accessible using proprietary web interfaces such as Google StreetView \cite{streetview}. Cesium Tiles was made available as plugins for state-of-the-art gaming engines including the Unreal Engine ($\sim$March 2021) and the Unity engine ($\sim$May 2022). As a result, simulation platforms built upon these technologies have newfound capabilities to use these highly realistic 3D models in simulations. 

Despite the benefits afforded by this new technology, a number of challenges exist to use this technology for robotic system development especially when developing and analyzing tracking, odometry, and mapping algorithms critical to this discipline. Challenges originate from how the geometric data is transmitted and rendered and other challenges originate from format and accessibility issues. 
 


The contributions of this article are:

\begin{itemize}
    \item a technical approach for controlling the Cesium Tiles plugin geometry caching to allow for large-scale 3D geometry analysis and sensing required for extracting volumetric (octree) models of the Cesium Tiles environment,
    \item an approach to reproduce a real-world experiment in simulation using the AirSim open-source simulator with the Cesium Tiles plugin,
    \item a quantitative and qualitative evaluation of how the algorithm results in the simulated environment compared with real-world experiments at the same location.
\end{itemize}

All of these contributions represent the first reporting of the impact that the Cesium Tiles technology has on the realistic simulation of robotic systems. Technical work with Microsoft AirSim \cite{airsim2017fsr} and the Cesium Tiles plugin resolves challenging technical issues that arise when using this simulation method. Experimental work examines the impact that this new expansive collection of 3D data can have on robotic simulation. Comparative simulation vs. real-world experiments provide initial insights into the effectiveness of these simulation methods for predicting real-world results from those that were simulated at specific real-world locations.

\section{Related Work}

The literature relating to this article is divided into two parts: one part that considers available aerial robotic simulation solutions and another which considers approaches for obtaining and applying high-fidelity 3D models of real-world locations.

\subsection{Aerial Simulation Solutions}

There are various simulation platforms for vehicles and environments catering to the diverse needs of researchers. Gazebo \cite{gazebo_paper}, with its open-source nature, stands as a versatile choice, emphasizing realism and adaptability. Agilicious \cite{agilicious} specializes in agile quadrotor flight, providing unique applications such as drone racing. RotorS \cite{rotors_sim}, integrated with the Robot Operating System (ROS), offers high-fidelity UAV simulation. Flightmare \cite{song2020flightmare}, part of the AirSim project, excels in simulating multiple drones for swarm robotics research. Kumar Robotics Autonomous Flight \cite{kumarRobotics1} addresses GPS-denied quadcopter autonomy. MIT's FlightGoggles \cite{flightgoggles} offers an immersive experience with photorealistic graphics. AirSim, developed by Microsoft, on top of the Unreal Engine, excels in generating highly realistic perceptual simulation data in complex and dynamic environments. 

\subsection{Digital Twins of Real-World Environment Models}

Digital twin technologies seek to create virtual models that replicate real-world contexts. Recent work being done using simulated real-world locations has been mainly done for advancing air mobility \cite{10311333,10124310,10311124}, while others generate environments to test different real-world constraints \cite{aerospace8050133,10298451}. Yet, these works use 2D satellite data and apply these textures to flat planar surfaces. Model 3D objects consist of building structures that are either hand-made or imported from the Open Street Maps database. 

This work represents a significant advancement over existing prior work in terms of 3D model realism for both geometry and appearance and allows heretofore unavailable flexibility in digital twin simulation by accessing the vast resources of Google's 3D map database. 


\begin{figure}
    \begin{subfigure}{0.45\linewidth}
        \centering
        \includegraphics[height=2.48cm]{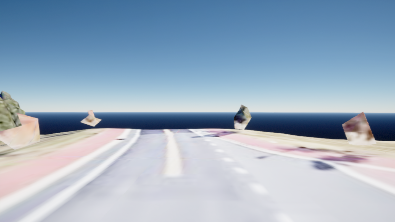}
        \caption{}
        \label{fig:cesium_bad_cache_sub}
    \end{subfigure}
    \begin{subfigure}{0.45\linewidth}
        \centering
        \includegraphics[height=2.48cm]{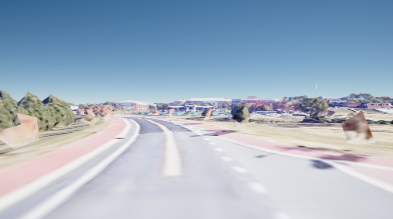}
        \caption{}
        \label{fig:cesium_good_cache_sub}
    \end{subfigure}
    \caption{(a,b) show how differences in the cine-camera  viewpoint and the viewpoints onboard perception sensors can generate incorrect image data. (a) shows a payload camera view distinct from the cine-camera; the black region on the horizon is due to a missing tile of model data. (b) uses our proposed method to tailor the cine-camera position and shows the missing geometry correctly rendered.}
    \label{fig:cesium_cache}
\end{figure}

\section{Methodology}

The methodology of this paper is broken into (3) parts:

\begin{enumerate}
    \item a method to indirectly control the tile caching to enable researchers to simultaneously load and render large 3D scenes using Cesium Tiles environments,
    \item a method to extract geometry and appearance measurements from Cesium Tiles environments, and
    \item an experimental evaluation of mapping algorithms in corresponding Cesium Tiles and real-world environments to analyze how simulated results for a specific location translate from simulation to reality.
\end{enumerate}

In this article, we apply a SLAM algorithm using a Cesium Tiles model of the UNC Charlotte campus football stadium and surrounding area and compare results with experimental results from the same algorithm generated from a real-world flight at the same location.

\subsection{Controlling Cesium Tiles Geometry Caching}
\label{sec:controlling_cinecamera}


The Cesium Tiles technology decomposes the geometric and appearance data compiled by Google into 3D blocks in a manner similar to games such as Minecraft. This computational architecture uses the viewpoint of the active camera, referred to as the \emph{cine-camera}, in its visible map regions to determine ``chunks" or, equivalently, ``tiles" from the map database to be transmitted to the client application's geometry cache and then rendered to the screen. Work in this article determines approaches that manipulate the viewpoint to allow geometric analysis of large geometric regions and simulation of sensor telemetry for sensors having omnidirectional capabilities.

A challenge with Cesium 3D tilesets is that only geometry tiles visible within the main view of the cinematic camera, referred to as the \emph{cine-camera}, are rendered. When the visible geometry inside the cine-camera view does not include the visible geometry of all other simulated sensors, the portions of the sensor data that reference data outside the view of the cine-camera will be missing, this causes problems for depth, RGB, and IR perceptual sensor data generation in the AirSim and can easily invalidate the simulated telemetry for these sensors. Figures \ref{fig:cesium_bad_cache_sub} and \ref{fig:cesium_good_cache_sub} show this problem and our proposed solution's result. In this case, the cine-camera is configured to have a top-down viewpoint whereas the camera sensor onboard the UAS has a forward-looking viewpoint. Cesium Tiles loads only geometry data tiles close to the vehicle to render the top-down scene and does not include the geometry viewed by the  UAS camera sensor. Figure \ref{fig:cesium_bad_cache_sub} shows a dark blue region where the missing tile data should be rendered. After moving the cine-camera to a distant location, the sensor visible data is a subset of the cine-camera visible region. Figure \ref{fig:cesium_good_cache_sub} shows an image from the same sensor after applying this modification to the cine-camera which fixes the problem.
 



\subsection{Extracting 3D Geometry from Cesium Tiles Environments}
\label{sec:gt_geometry}

\begin{figure}
    \centering
    \includegraphics[width=0.95\linewidth]{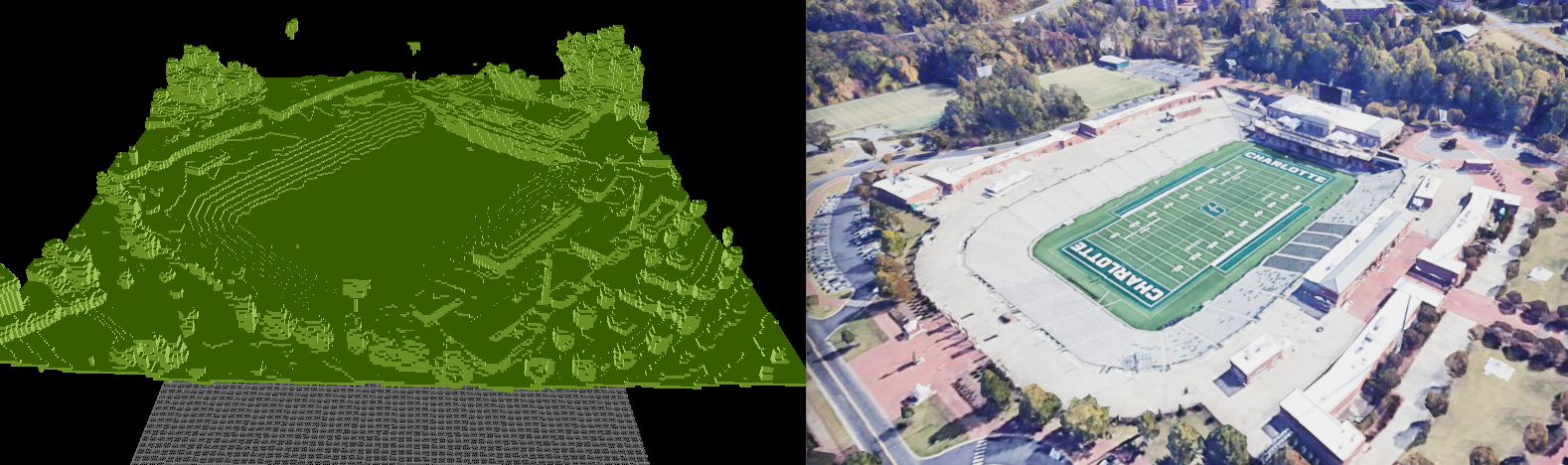}
    \label{fig:geometry_to_voxels}
    \caption{An example of converting a geometric model to a voxel model using AirSim's \emph{simCreateVoxelGrid()} function.}
\end{figure}

While there are clear and direct paths for converting 3D geometry representations to the Cesium Tile format, it does not appear possible to convert data in this format to an open-source format. This presents a challenge for algorithms that require exact knowledge of the scene geometry. In this work, we seek to evaluate mapping and odometry results for SLAM algorithms and this technical challenge prevents this evaluation since the environment geometry is not directly available.

A geometric representation of the environment is made available using AirSim's built-in \emph{simCreateVoxelGrid()} function. This function accesses the currently loaded world geometry and computes a voxel occupancy grid sampling from the existing geometry. The function input parameters are the grid center position, grid dimensions, and voxel cell resolution. The voxel model returned is a coarse representation of the underlying polygonal surface mesh and can be made arbitrarily more accurate with small enough resolution values at the expense of size/memory resources. 

Our experimental results combine the viewpoint manipulation method previously described in combination with this geometry extraction approach to extract large-scale geometric models from Cesium Tile models.

The voxel grid is post-processed to convert the occupancy grid to a point cloud approximation of the model surface. AirSim stores voxel grid geometries in \emph{binvox}. The point cloud is computed from the \emph{binvox} representation by converting each occupied cell into a 3D $(x_n,y_n,z_n)$ point measurement as described in equation (\ref{eq:voxel_coord}) where $(i,j,k)$ is the voxel index starting at $(0,0,0)$ and $d$ is the grid dimension.

\begin{equation}
    (x_n, y_n, z_n) = \left(\frac{i + 0.5}{d}, \frac{j + 0.5}{d}, \frac{k + 0.5}{d} \right)
    \label{eq:voxel_coord}
\end{equation}




\subsection{Mapping Algorithm Evaluation Metrics}

\begin{figure}
    \centering
    \includegraphics[width=0.8\linewidth]{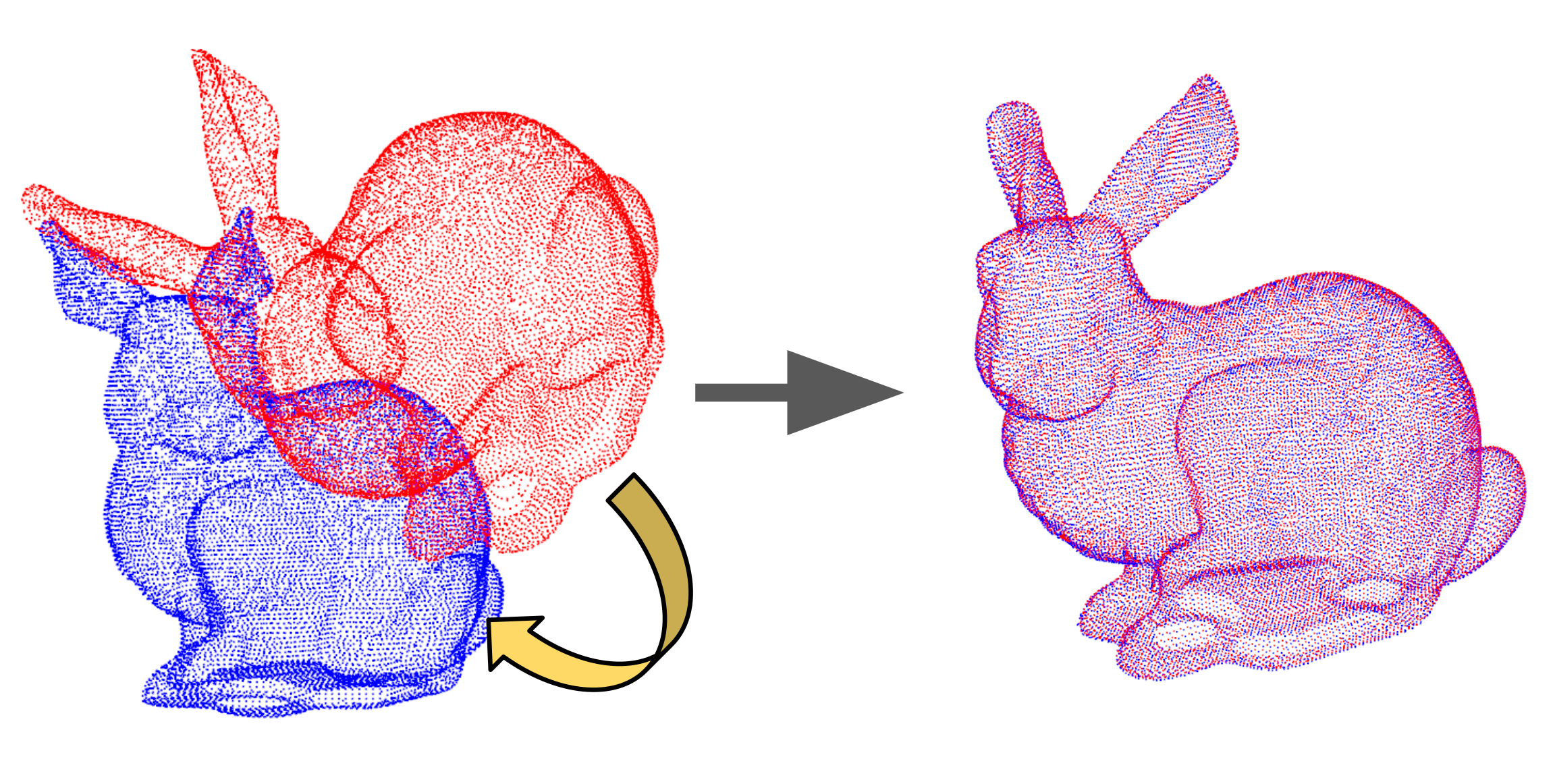}
    \caption{An example of using ICP algorithm to align two point clouds \cite{Zodage-2021-129203}. Red: source point cloud. Blue: target point cloud. Purple: registration result.}
    \label{fig:icp_example}
\end{figure}

The geometric models extracted from Cesium can be leveraged by researchers to evaluate the geometric accuracy of 3D maps from real-world SLAM experiments. To do this, a model of the real-life experiment location is requested using the Cesium Tiles plugin, and the experimental data is aligned to the Cesium Tiles model. While a coarse alignment can be achieved using GPS coordinates, highly accurate alignments are possible using point cloud alignment algorithms.

Figure \ref{fig:icp_example} demonstrates the concept of point cloud alignment where two point clouds (blue, red) are measured in distinct coordinate frames. The Iterative Closest Point (ICP) algorithm \cite{besl1992method} is employed to estimate the coordinate transformation that aligns the coordinate frame of the moving (red) dataset to the coordinate frame of the fixed (blue) dataset. 

This work evaluates mapping performance by applying the ICP algorithm to align mapping data with Cesium Tiles environmental models. After alignment, correspondences are defined between the 3D points of the map data and the environmental model by associating each 3D point from the mapping data to the closest point in the environment model. Since point clouds often include many outlier measurements, correspondences are typically only computed for points that lie within a user-specified threshold distance.
 

This work uses the methods of section \ref{sec:gt_geometry} to define the environment model and uses this model as ground truth. Experimentally estimated map data point clouds are registered to the ground truth geometry for analysis of geometric accuracy. Once the point correspondences are established, the geometric accuracy is evaluated by measuring the distances between these corresponding points. Key metrics such as the mean point distance and the standard deviation are then calculated to quantitatively assess the spatial fidelity and alignment precision of the reconstructed point cloud with respect to the ground truth model. 







\section{Results}

This section focuses on the data collection from a real-world flight and simulated flight, 3D point cloud reconstruction of the environment, and finally registration of the point clouds to assess the accuracy and quality of the reconstruction.

\subsection{Real-World Experiment and Reproduction in Simulation}

The location of both the real-world flight and the simulated flight was the Jerry Richardson Stadium at the University of North Carolina at Charlotte having (latitude, longitude) coordinates $(35.310344, -80.740007)$.

\begin{figure}
    \begin{subfigure}{0.45\linewidth}
        \centering
        \includegraphics[height=2.6cm]{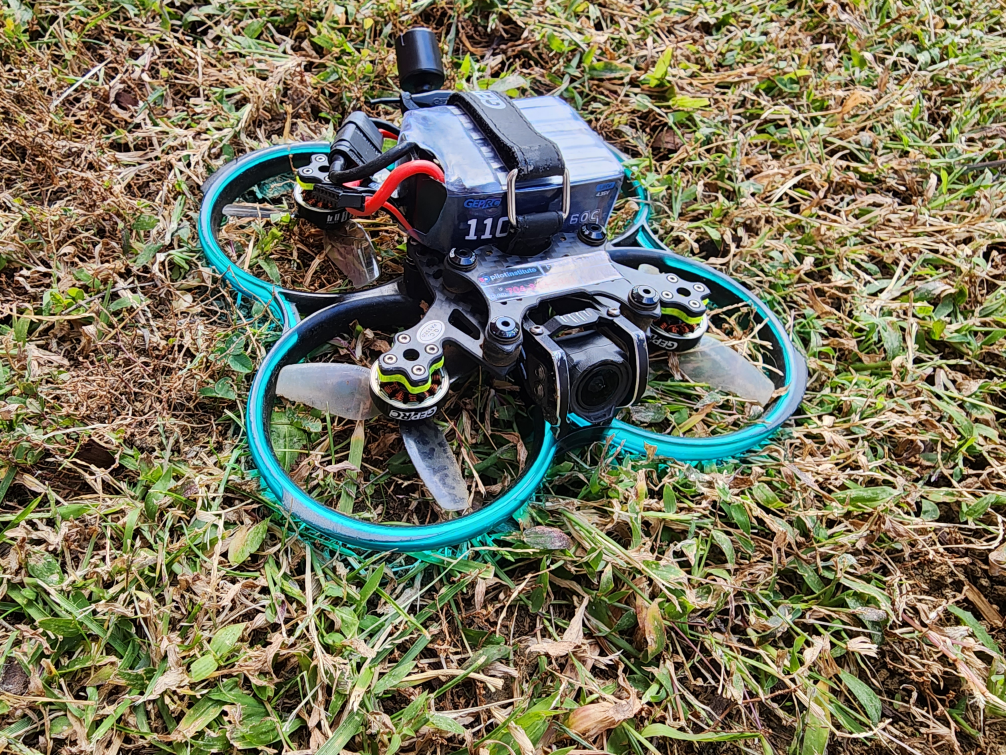}
        \caption{}
        \label{fig:dji_drone}
    \end{subfigure}
    \begin{subfigure}{0.45\linewidth}
        \centering
        \includegraphics[height=2.6cm]{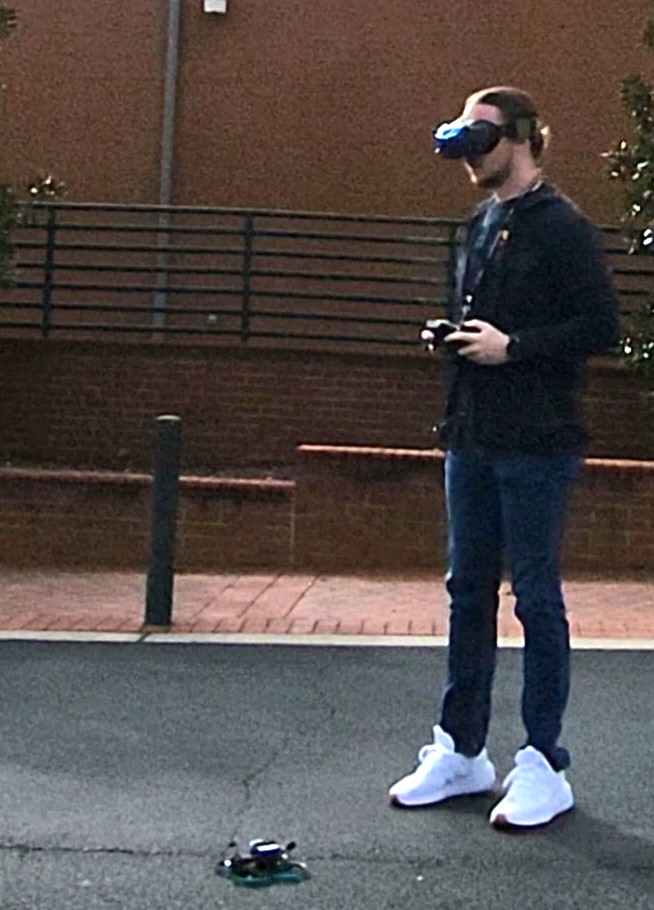}
        \caption{}
        \label{fig:dji_field_experiment}
    \end{subfigure}
    \caption{(a) Cinebot 30 drone with DJI O3 camera. (b) Image of the drone operator during a flight test.}
    \label{fig:dji_drone_camera}
\end{figure}

\subsubsection{Real-world Data Collection}
\label{sec:real_flight}

The real-world data was collected on a GEPRC Cinebot 30 drone equipped with a DJI O3 Air Unit for capturing video, as seen in Fig. \ref{fig:dji_drone_camera}. The real-world flight was piloted with first-person view (FPV) goggles and lasted for 3 minutes. The first 33 seconds were used as data for analysis. This excerpt of the data includes views of rich geometric structures like buildings and trees. Cinebot image sensor data was also down-sampled from 120 fps to 30 fps to reduce data redundancy.  

\subsubsection{Simulated Reproduction of the Real-World Experiment}

Simulated experiments were conducted to reproduce the real-life vehicle, sensor, and trajectory using a virtual environment model of the UNC Charlotte football stadium generated with Unreal Engine and the Cesium plug-in. As shown in Fig.~\ref{fig:flight_trajectory} the real-life flight trajectory and simulated trajectory have differences but the selected excerpts of recorded image telemetry are similar. 



The simulated flight lasted for 2 minutes and 51 seconds. As in the real-world experiment, an excerpt of the flight data was extracted where the virtual sensor exhibited similar viewpoints of similar structures for comparative analysis.

The specifics of the sensor configuration for real-life and simulated experiments are shown in Table \ref{tab:experiment_attributes}. The simulated and real-world camera sensors shared the same Field Of View (FOV) and resolution parameters but different frame rates. A real-world and simulated experimental RGB image is shown in Figs. \ref{fig:real_world_flight_segment} and \ref{fig:sim_world_flight_segment}, respectively.


\begin{table}
    \centering
    \begin{tabular}{|c|c|c|}
        \hline
        &  Simulated & Real-world \\ [0.5ex]
        \hline \hline
        Field of View (deg) & 155 & 155 \\
        \hline
        Resolution (HxV pixels) & 2688x1512 & 2688x1512 \\
        \hline
        Framerate (fps) & 5 & 30 \\
        \hline
        Duration (frames) & 342 & 1000 \\
        \hline
    \end{tabular}
    \caption{Real-world experimental sensing parameters are duplicated to the AirSim simulated environment.}
    \label{tab:experiment_attributes}
\end{table}


\subsubsection{Algorithmic Data Generation in Simulation and Real World Experiments}

Point cloud maps were generated from each flight using a state-of-the-art mapping algorithm, Direct Sparse Odometry (DSO) \cite{engel2017direct}. DSO is a visual odometry technique that adapts Structure from Motion (SfM) methods for 3D reconstruction. It directly estimates the camera motion and the sparse 3D structure of the environment from a sequence of 2D images by minimizing photometric errors. 

\begin{figure}
    \begin{subfigure}{0.45\linewidth}
        \centering
        \fbox{\includegraphics[height=2.34cm]{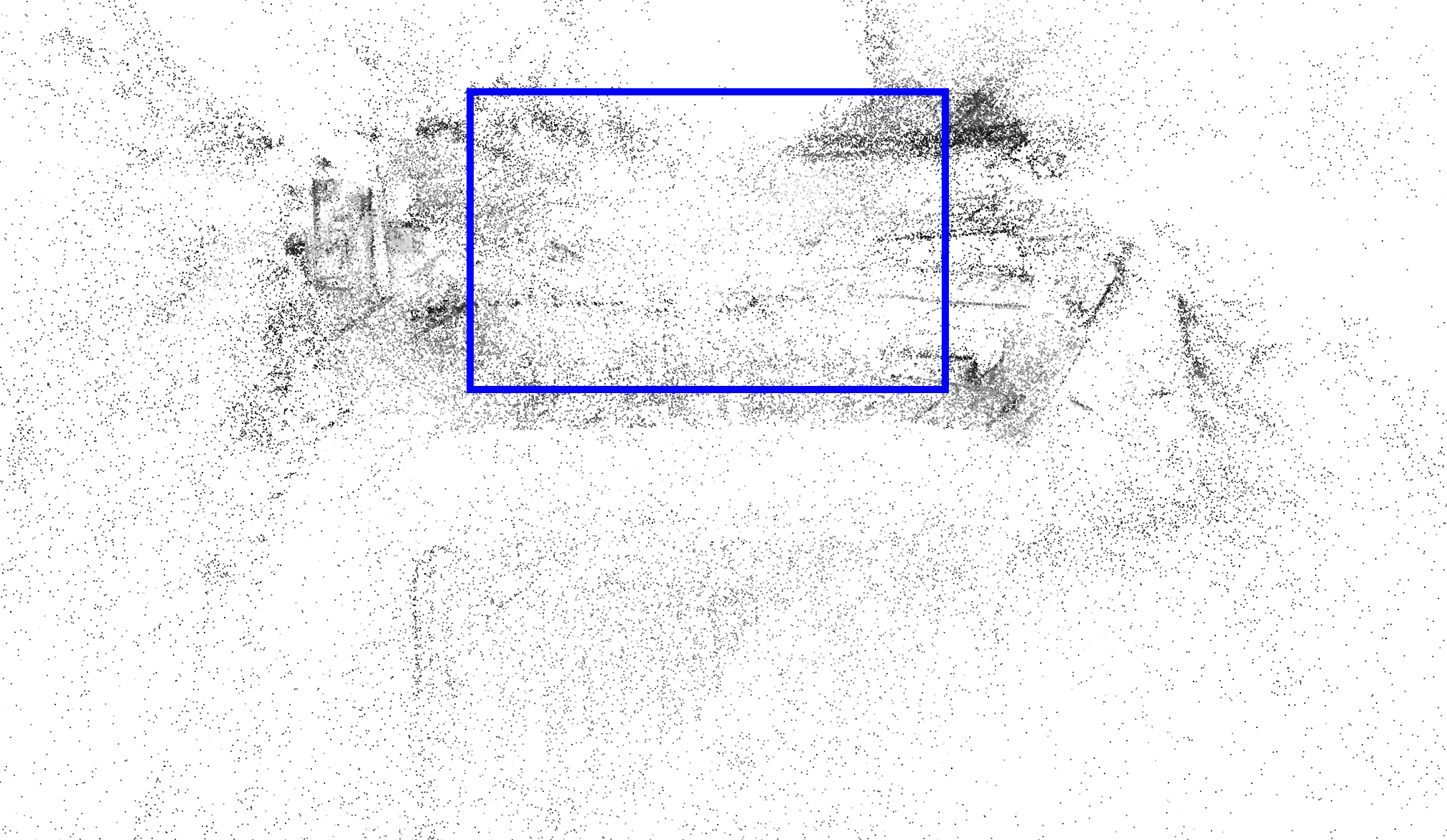}}
        \caption{}
        \label{fig:dji_point_cloud}
    \end{subfigure}
    \begin{subfigure}{0.45\linewidth}
        \centering
        \fbox{\includegraphics[height=2.34cm]{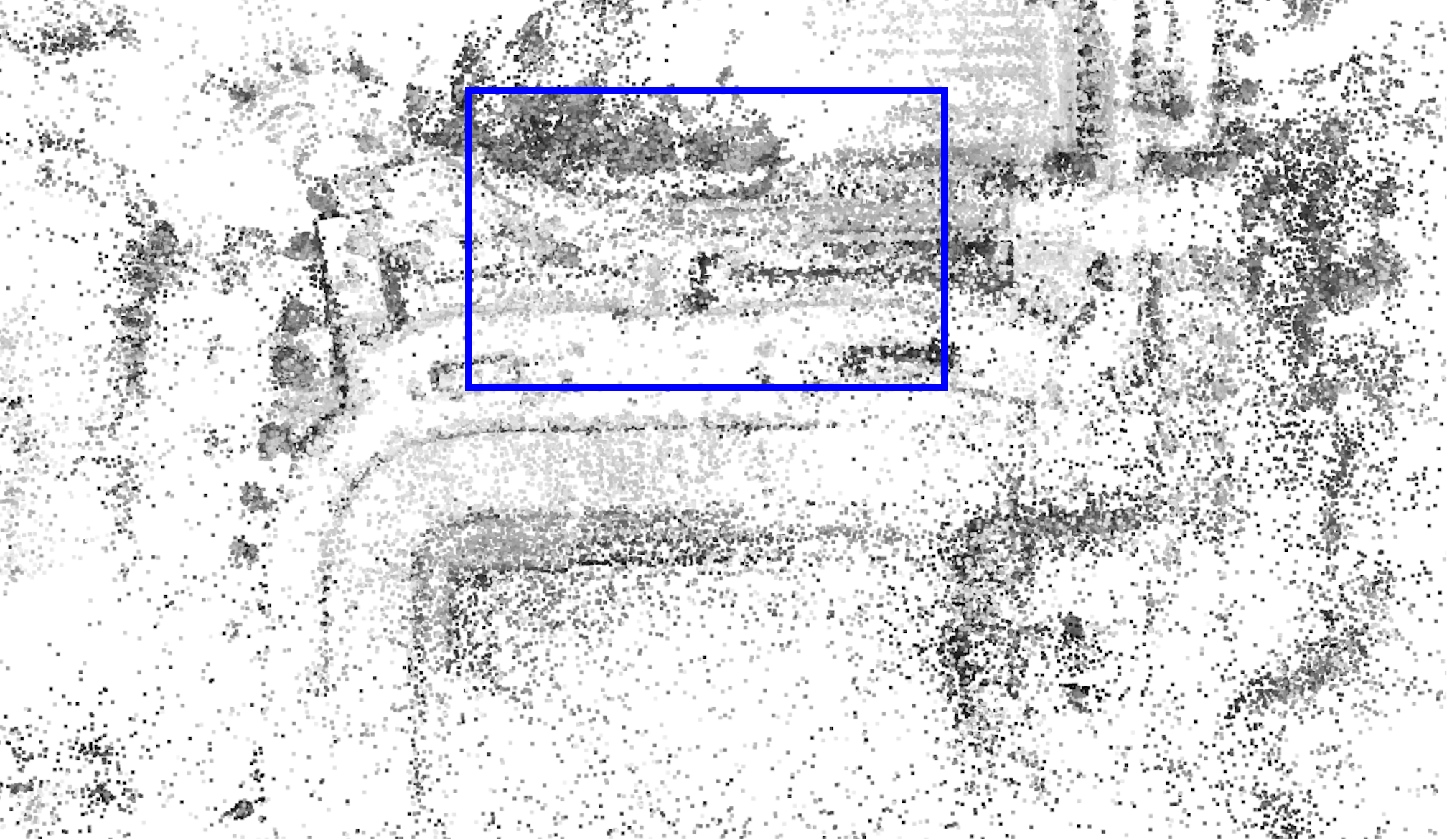}}
        \caption{}
        \label{fig:airsim_point_cloud}
    \end{subfigure}
    \\
    \begin{subfigure}{0.45\linewidth}
        \centering
        \includegraphics[height=3.2cm]{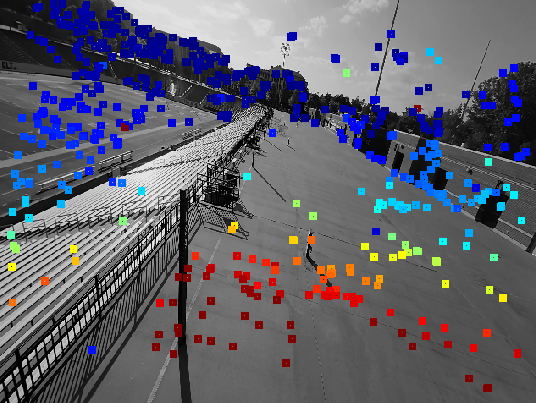}
        \caption{}
        \label{fig:dji_frame}
    \end{subfigure}
    \begin{subfigure}{0.45\linewidth}
        \centering
        \includegraphics[height=3.2cm]{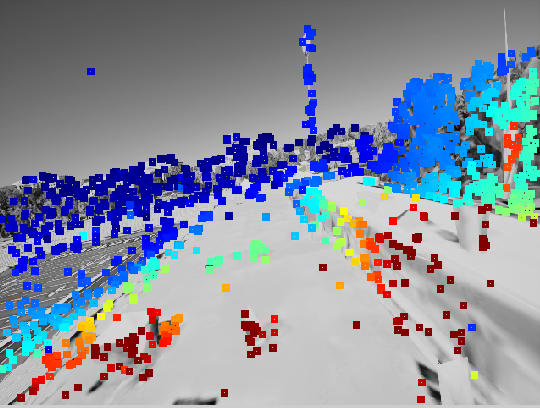}
        \caption{}
        \label{fig:airsim_frame}
    \end{subfigure}
    \caption{(a, b) show algorithm mapping data generated from the real-life stadium and its virtual model using Cesium Tiles. The blue boxes indicate the regions reconstructed from virtual data while missing in the real reconstruction. (c, d) show frames from real and simulated image data with the color points indicating the feature points detected by DSO. The two frames also correspond to the area in the blue box.}
    \label{fig:point_clouds}
\end{figure}

Figures~\ref{fig:dji_point_cloud} and  \ref{fig:airsim_point_cloud} show point clouds reconstructed by DSO from the simulated flight and the real flight respectively. Both of the point clouds include salient large-scale structures such as the wall that separates the football field from the stands, the concession buildings, and the stadium seating. 



\subsection{Extracting Large Scale Models from Cesium Tiles}

\begin{figure}
    \centering
    \includegraphics[width=0.65\linewidth]{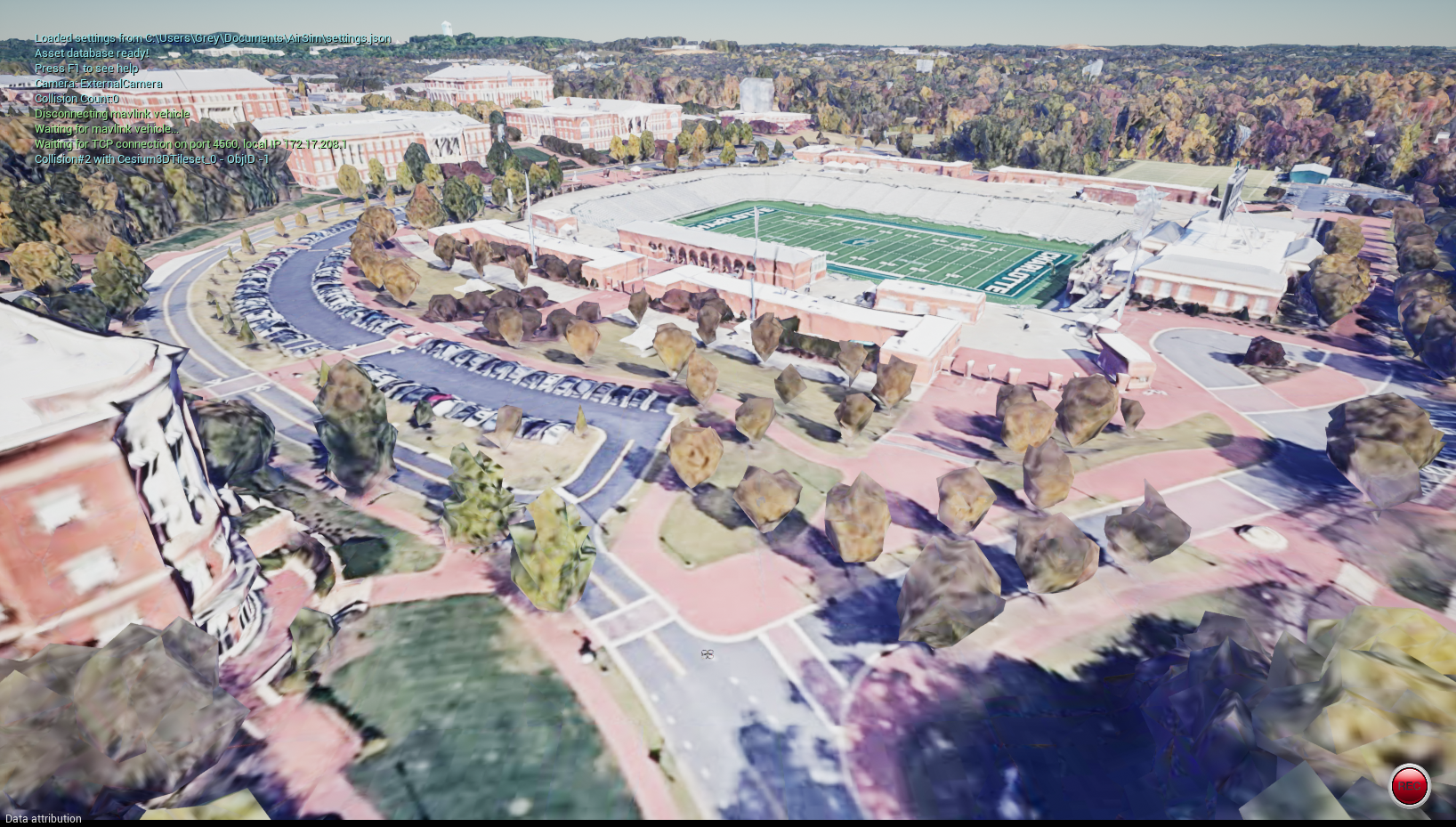}
    \caption{A high altitude viewpoint of the UNC Charlotte campus and football stadium. Cine-camera viewpoints from high altitudes compel Cesium Tiles to load and render the entire 3D region of interest simultaneously.}
    \label{fig:cesium_unreal_example}
\end{figure}

Figure \ref{fig:cesium_unreal_example} shows the position manually set for the cine-camera view to provide reliable perceptual sensor telemetry using the method of section \ref{sec:controlling_cinecamera}. The viewpoint shown compels the Cesium plugin to transfer, load, and render all geometry of interest for analysis in the article. As described in the methodology, the cine-camera must be carefully positioned to reliably extract large-scale geometric models of the environment using the Cesium Tiles plugin. 

\begin{figure}
    \centering
    \includegraphics[width=0.6\linewidth]{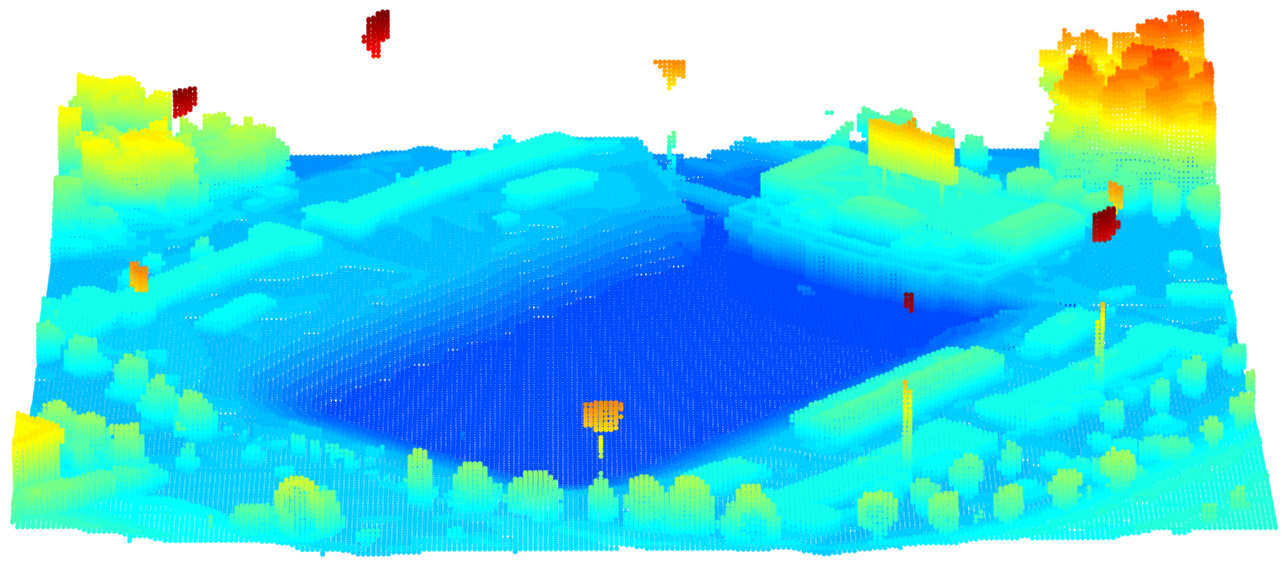}
    \caption{A 250m$\times$250m point cloud with 1m resolution  of the UNC Charlotte football stadium generated from a voxel model.}
    \label{fig:voxel_point_cloud}
\end{figure}

Figure \ref{fig:voxel_point_cloud} shows the voxel point cloud extracted via the \emph{simCreateVoxelGrid()} with the cine-camera configured as shown in Fig. \ref{fig:cesium_unreal_example}. This function extracts geometric data available from a low-level function called the Unreal Engine which provides an occupancy map of the scene. The function arguments require a seed $(x,y,z)$ position, a specification of the $(x,y,z)$ dimension of a voxel, and a $(x,y,z)$ resolution which jointly determines the 3D range and resolution of the returned volumetric scene model. The voxel model generated consisted of 158,446 3D $(x,y,z)$ measurements which were processed according to the steps described in the methodology.

\subsection{Quantitative Evaluation of Mapping Algorithm Results}

\begin{figure}
    \centering
    \includegraphics[width=0.6\linewidth]{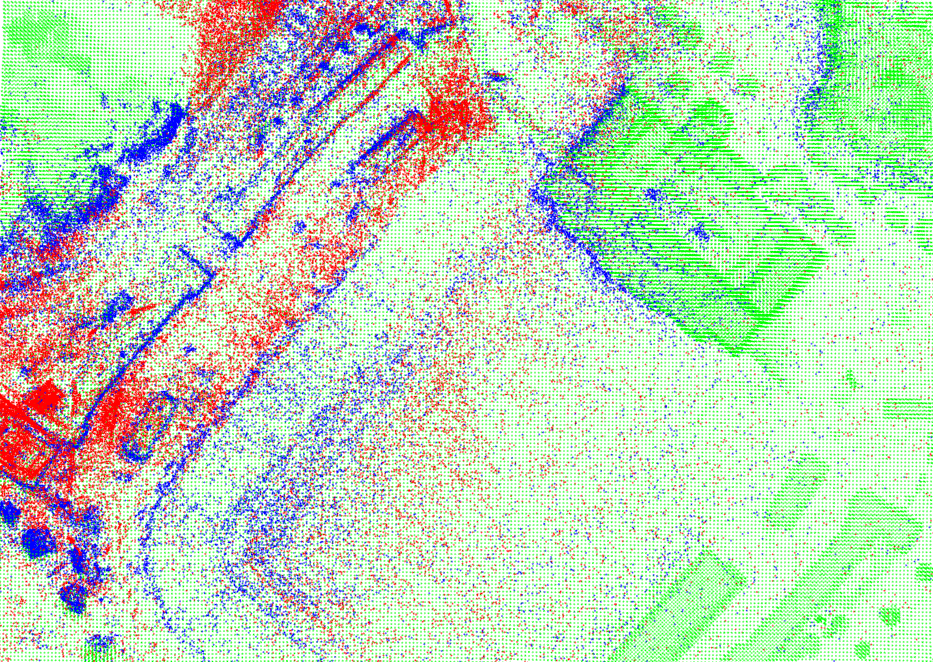}
    \caption{Top-down view of the combined point cloud with the voxel point cloud in green, simulated point cloud in blue, and real-world point cloud in red. Qualitatively the algorithm results for the simulated Cesium Tiles environment and the real-world experiments at the same location appear similar.}
    \label{fig:combined_point_cloud}
\end{figure}

\begin{table}
    \centering
    \begin{tabular}{|c|c|c|c|}
        \hline
        &  Simulated & Real-world \\ [0.5ex]
        \hline \hline
        Algorithm Map Points Estimated (\# pts) & 84,029 & 71,918 \\
        \hline
        Model-to-Map Correspondences (\# pts) & 36,178 & 27,820 \\
        \hline
        Mean error (m) & 0.5548 & 0.5733 \\
        \hline
        Std. Dev. (m) & 0.1885 & 0.1998 \\
        \hline
    \end{tabular}
    \caption{Quantitative statistics comparing DSO mapping algorithm results in simulation and real-world experiments after alignment with a Cesium Tiles environment model from the experimental location.}
    \label{tab:error_table}
\end{table}

\begin{figure}
    \centering
    \includegraphics[width=0.7\linewidth]{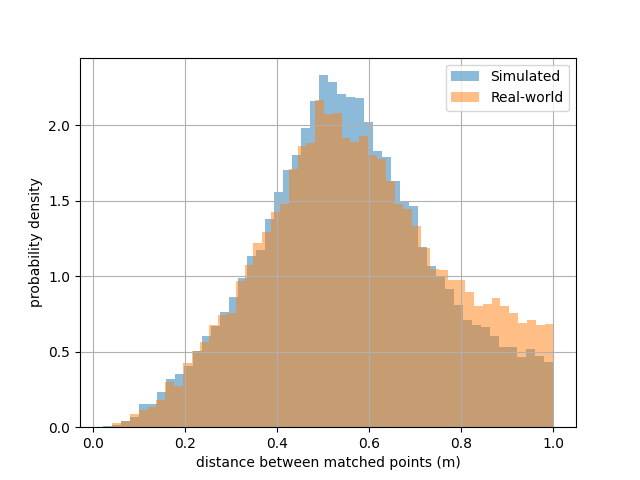}
    \caption{Quantitive analysis of the distribution of the closest point distances for correspondences from each simulation point cloud to voxel point cloud. The mean correspondence errors in simulation and real experiments are close in value while the variance in simulation is smaller than that in real experiments. }
    \label{fig:probabilityDistribution}
\end{figure}

All collected point clouds were registered to the voxel point cloud using the ICP algorithm.  To align the point clouds from Fig.~\ref{fig:airsim_point_cloud} and Fig.~\ref{fig:dji_point_cloud} with the extracted voxel geometry shown in Fig. \ref{fig:voxel_point_cloud} subsets of the complete data sets were used. Alignment work seeks to use corresponding regions from the experimentally generated point cloud data to perform alignment. This promises to generate a similar quality of alignment for both real-world and simulated alignment results since similar surface measurements from the data are being used.

The specific collections of regions selected for alignment included the corner of the buildings and the corner of the football field. The selected regions for each experimentally generated point cloud were then used as input to the ICP algorithm to align the voxel geometry to both of the experimental point cloud data sets. The configuration used for the ICP algorithm is a search radius of 1 meter, root mean square error threshold of 0.00001, and maximum iteration time of 1500.

After alignment, a global correspondence is calculated between each point cloud and the voxel point set. Correspondences are created for each $(x,y,z)$ location in the point cloud that is within 1 m. of a point in the voxel point set.
Figure \ref{fig:combined_point_cloud} shows the registered point clouds together where the green points are the voxel point cloud, the blue points are the simulated point cloud, and the red points are the real-world point cloud.

Table \ref{tab:error_table} provides statistical values, including the number of correspondences, mean, and variance of correspondence distances between the voxel point cloud and point cloud data sets. The simulated data's point cloud contains 14.4\% more points than the real-world reconstruction, as evident in Fig. \ref{fig:airsim_point_cloud} (simulated) and Fig.~\ref{fig:dji_point_cloud} (real-world), showcasing denser simulated data. The amount of correspondence reflects the same difference. Despite fewer frames in the simulation data as shown in Table \ref{tab:experiment_attributes}, such a difference in the point cloud set is caused by more feature points detected by DSO in the simulation data. An example of this behavior is shown in Fig. \ref{fig:airsim_frame} and Fig.~\ref{fig:dji_frame} which correspond to the reconstruction area in the blue boxes in point cloud figures. The real-world image data quality suffers from changes in lighting conditions which cause pixel intensity inconsistencies for the same scene point seen by different frames. Such inconsistency challenges DSO to successfully track and reconstruct these points. The simulated environment, however, does not have this issue. The mean error for both the simulated and real-world point clouds is relatively close to each other while the variance in the simulation point set is 5.9\% smaller than that in real experiments, which is also reflected by Fig.~\ref{fig:probabilityDistribution}, the histogram of the fitting error between the two flights. 

\begin{figure}
    \centering
    \includegraphics[width=0.75\linewidth]{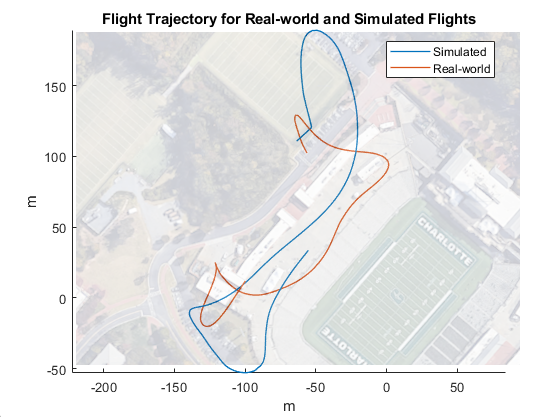}
    \caption{Top-down view of real-life (red) and virtual (blue) flight trajectories after point cloud registration.}
    \label{fig:flight_trajectory}
\end{figure}

Fig. \ref{fig:flight_trajectory} shows a top-down view of the flight trajectories for both the simulated and real-world flight using the alignments resulting from point cloud registration. 

\subsection{Discussion}

Results that compare real-world DSO algorithm data with the Cesium Tiles environment model indicate that real-world experiments can be conducted and compared against Cesium Tiles environments for effective SLAM performance analysis. This is evidenced by error distributions for SLAM data that agree in distribution with results reported in prior work. This promises to significantly impact real-world experiments where a detailed model of the environment is not available.

Simulation work that produces a similar sequence of telemetry to the real-world experiment exhibits very similar statistical metrics. As shown in Table \ref{tab:error_table} the simulated 3D map point cloud and corresponding real-life point cloud have similar metrics in terms of the number of points and, to a lesser degree, the number of correspondences (using a threshold of 1m.). Figure \ref{fig:probabilityDistribution} shows the distributions of the distances between ground truth map points and their estimates in the real-lift (orange) and simulated (blue) contexts. These two distributions are highly correlated having a 1.9cm difference between their mean values and a 1.1cm difference in standard deviation. One explanation for the differences in distribution is the reduced resolution available for the textures that provide the appearance data for the Cesium Tile environment model. This manifests in the simulated sensor images as blurry imagery which may reduce the noise in the associated point cloud data.

\section{Conclusion}

This article evaluates the application of digital twin environments made available by accessing the Google 3D map database made available in the AirSim simulator using the Cesium Tiles plugin that has been recently made available. The results of this analysis indicate that researchers can effectively use these environment models to evaluate the performance of SLAM algorithms conducted at locations where the ground truth model of the scene geometry is not available. Further work triaged several technical challenges associated with the Cesium plugin to provide a capability of approximately reproducing a real-life experiment in the Cesium virtual environment by traversing a similar path and collecting a virtually simulated version of the experiment telemetry. The effectiveness of this experimental reproduction method was evaluated by running a SLAM algorithm on real-life and virtual experimental telemetry and performing a comparative analysis of the 3D SLAM point cloud results. Technical issues associated with this analysis were overcome via the careful application of point cloud alignment algorithms for the point cloud data. Findings indicate that the algorithm results on real-life and on simulated data exhibit a high degree of similarity. This indicates that use of Cesium Tiles environments as a virtual digital twin for real-life experiments will provide representative results for such algorithms. The impact of this can be significant, potentially allowing expansive virtual testing of robotic systems at specific deployment locations to develop solutions that are tailored to the environment and potentially outperforming solutions meant to work in completely generic environments.



\bibliographystyle{IEEEtran}
\bibliography{main}

\end{document}